\documentclass{bmvc2k}

\usepackage{floatrow}
\title{Built-in Elastic Transformations for Improved Robustness}

\addauthor{Sadaf Gulshad*}{s.gulshad@uva.nl}{0}
\addauthor{Ivan Sosnovik*}{i.sosnovik@uva.nl}{0}
\addauthor{Arnold Smeulders}{a.w.m.smeulders@uva.nl}{0}

\addinstitution{
UvA-Bosch Delta Lab \\ University of Amsterdam \\The Netherlands
}

\runninghead{Gulshad, Sosnovik, and Smeulders}{Built-in Elastic Transformations.}

\def\etal{\emph{et al}\bmvaOneDot}

\begin{document}

\maketitle

\begin{abstract}
We focus on building robustness in the convolutions of neural visual classifiers, especially against natural perturbations like elastic deformations, occlusions and Gaussian noise. Existing CNNs show outstanding performance on clean images, but fail to tackle naturally occurring perturbations. In this paper, we start from elastic perturbations, which approximate (local) view-point changes of the object. We present elastically-augmented convolutions (EAConv) by parameterizing filters as a combination of fixed elastically-perturbed bases functions and trainable weights for the purpose of integrating unseen viewpoints in the CNN. We show on CIFAR-10 and STL-10 datasets that the general robustness of our method on unseen occlusion, zoom, rotation, image cut and Gaussian perturbations improves, while significantly improving the performance on clean images without any data augmentation. 
\end{abstract}

%-------------------------------------------------------------------------
\section{Introduction}
\label{sec:intro}
When designing real-world neural network classifiers, models need to be robust to input transformations and perturbations. Current architectures tend to fail even with small changes in the input by adding blur or noise to the image \cite{DBLP:journals/corr/abs-1806-00451,azulay2018deep,dodge2017study}. Robustifying neural networks against such and other natural perturbations is important before deployment in practice. 

Robey \etal \cite{robey2020model} proposed to train the classifiers on naturally perturbed images generated using generative models. Rusak \etal \cite{rusak2020simple} showed that training a network on properly tuned Gaussian or speckle noise enhances its generalization to other perturbations as well. Adversarial training for robustification Good Fellow \etal \cite{goodfellow2014explaining} demonstrated less bias to texture in Zhang \etal \cite{zhang2019interpreting}. At the same time, Engstrom \etal \cite{engstrom2019exploring} and Gulshad \etal \cite{gulshad2021natural} showed that adversarial training does not generalize well to natural perturbations like rotations, translations, occlusions and blur. These approaches are based on data augmentation during training. In this work, instead of creating training images with different deformations, we aim our attention to the network architecture directly. The strategic advantages of implementing perturbations in the network are three-fold: there is no need to change the data, implementation in the network permits future optimization of the computations, and the network transformation permits for mathematical guarantees.

We take the constraint not to increase the number of images in learning; we assume no more images are available nor are there perturbed versions of images. We aim to get better classification performance from the given training images. We start from the observation that in practice frequent and important deformations occur in the image when the camera changes its viewing angle, when the lighting is dark and hence the image is noisy, and when the object is partially occluded. In all such circumstances, elastic transforms will provide a better variety in the input of the next layer and hence better descriptors for differentiating between two images without adding new images to learn from. As a by product, we demonstrate that these elastic transforms enhance the performance on clean images significantly. 

Global elastic transforms approximate view point changes, while local transformations cover out-of-plane rotations of the object. In medical images, the shape and volume of organs may vary Wang \etal \cite{wang2001elastic} and Buslaev \etal \cite{buslaev2020albumentations}, while in real-world views birds may be rotating relative to the camera. The same happens when the object sits still and the scene is dynamic as in ocean waves, or when  the camera moves as in action recognition and pose estimation. We observe that local elastic transforms provide a good approximation of many practical local variations in the image space one wants to be invariant under.

\begin{figure*}
    \centering
    \includegraphics[width=\linewidth]{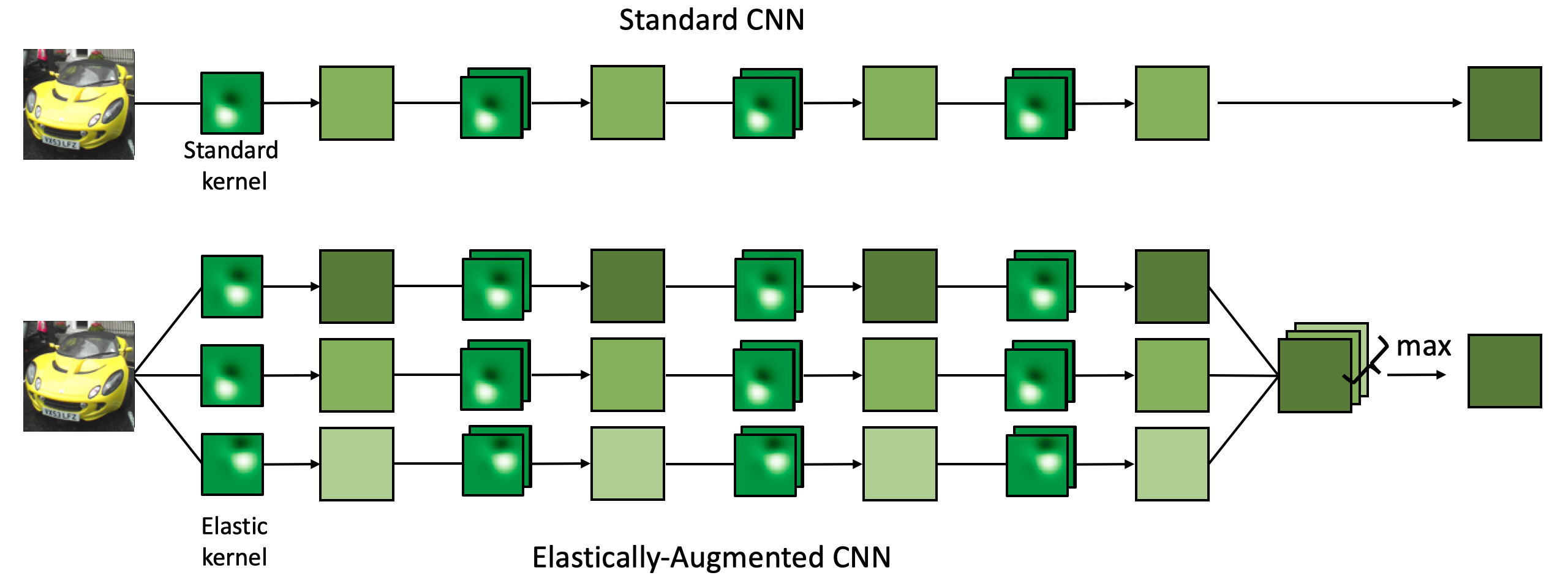}
    \caption{Top: A standard CNN with 4 convolutional layers. Bottom: Its elastically-augmented (EAConv) variant. By multiplying the fixed basis with the trainable weights, a single network is transformed in to a network with multiple paths, each path with a different basis. At the end, the maximum is selected. }
    \label{fig:Architecture}
\end{figure*}

In this work, we incorporate elastic transformations in convolutional neural networks and propose \textit{Elastically-Augmented Convolutions}. We do so by defining elastic transformations \textit{a priori} and learning the weights of the kernels. Jaderberg \etal \cite{jaderberg2015spatial}, Felzenszwalb \etal \cite{felzenszwalb2009object} and Dai \etal \cite{dai2017deformable}  focused on integrating similar transformations in the network and learning their parameters. However, the aim of previous approaches is to combat the transformations, while here our aim is  robustification for unseen transforms, with the following contributions: 
\begin{itemize}
    \item We propose the theory for elastically-augmented convolutional neural networks. \vspace{-2mm}
    \item We introduce \textit{Elastically-Augmented Convolutions} to integrate unseen viewpoints in the convolutional neural networks for enhancing their general robustness. \vspace{-2mm}
    \item We demonstrate that by incorporating elastic variations in the convolutions of the network we improve the performance on clean images, leading to the state of the art on STL-10 dataset i.e. $94.48$, and CIFAR-10 i.e. $94.50$ (without any data augmentation). \vspace{-2mm}
    \item We demonstrate \textit{specific} robustness for elastic transforms and, remarkably, \textit{general} robustness for Gaussian, occlusion, rotation, cut and zoom perturbations unseen during training.  \vspace{-2mm}
\end{itemize}

% \vspace{5mm}
\section{Related Work}

\paragraph{Robustness to Natural Perturbations}
A vast majority of work in the robustness of computer vision focuses on robustification against carefully designed perturbations, i.e. \textit{adversarial perturbations} \cite{madry2017towards,wong2018provable,madaan2019adversarial,goodfellow2014explaining}. However, adversarial robustification can not capture naturally occurring perturbations, e.g. rotations, translations, blur as was demonstrated in Engstrom \etal \cite{engstrom2019exploring} and Gulshad \etal \cite{gulshad2021natural}. Furthermore, a trade-off is also observed between robustness and clean image accuracy when networks are robustified with
adversarial training \cite{zhang2019interpreting,zhang2019theoretically,tsipras2018robustness}.

To improve the robustness against natural perturbations, Schneider \etal \cite{schneider2020improving} proposed to use batch normalization performed on perturbed images instead of clean ones. Similarly, Tang \etal \cite{tang2021selfnorm} introduced two different normalization techniques, Selfnorm and Crossnorm to enhance the robustness against perturbations. Benz \etal \cite{benz2021revisiting} also utilized perturbed samples and proposed to rectify batch normalization statistics for enhancing the robustness of neural networks against perturbations. 
Simultaneously, Rusak \etal \cite{rusak2020simple} introduced a noise generator that learns uncorrelated noise distributions. Training on these noisy images enhanced the performance against natural perturbations. Gulshad \etal \cite{gulshad2021natural} trained on images with adversarial as well as natural perturbations like occlusions or elastic deformations, while achieving good generalization for many other unseen perturbations. Robey \etal \cite{robey2020model} and Wong \etal \cite{wong2020learning} argued that it is impossible to capture all possible natural perturbations mathematically. Therefore, they used generative models to generate images with perturbations to train the network. 

Instead of training with perturbed inputs, in this work we integrate predefined common perturbations into the network to enhance robustness.

\paragraph{Built-in Image Transformations}
Initially, geometric transformations were modeled in the neural networks by small units that locally transformed their inputs for modeling geometric changes, i.e. capsules  \cite{hinton1981parallel}. Later, Jaderberg \etal \cite{jaderberg2015spatial} introduced a transformer module in the network to wrap feature maps by global transformations. However, learning the parameters of the transformations introduced by Jaderberg \etal \cite{jaderberg2015spatial} is known to be difficult and computationally expensive. In  similar spirit, Felzenszwalb \etal \cite{felzenszwalb2009object} and Dai \etal \cite{dai2017deformable} focused on integrating spatial deformations in CNNs. Both methods require large datasets for learning, while our aim is to learn from small datasets and generalize the performance to include perturbations on images never seen before.

\section{Method}

\subsection{Image Transformations}
Consider an image $f$. It can be reshaped as a vector $\mathbf{f}$. A wide range of image transformations can be parametrized by a linear operator: scaling, in-plane rotations, shearing. Other transformations, such as out-of-plane rotations, can not be parametrized in an image agnostic way. However, for small deviation from the original image Taylor expansions can be used, which gives a linear approximation for many image transformations of practical use. Indeed,
\begin{equation}
\label{eq:taylor}
    T[f](\epsilon)
    \approx T[f](0) + \epsilon\left.\Big(\frac{\partial T[f]}{\partial \epsilon}\Big)\right\rvert_{\epsilon=0}
    % = \mathbf{f}  + \epsilon\left.\Big(\frac{\partial T[f]}{\partial \epsilon}\Big)\right\rvert_{\epsilon=0} 
    = \mathbf{f}  + \epsilon\mathbf{L}_T\times \mathbf{f}
    = (\mathbf{I} + \epsilon\mathbf{L}_T)\times \mathbf{f} 
    = \mathbf{T} \times \mathbf{f}
\end{equation}
where $T$ is a transformation, $\epsilon$ is the parameter of the transformation and $\mathbf{T}$ is a linear approximation of $T$ for small values of the parameter. For scaling the parameter is the logarithm of the scaling factor, for rotations it is the angle, and so on. $\mathbf{L}_T$ is a matrix representation of an infinitesimal generator of $T$. 

An image $f$ can also be viewed as a real-value function of its coordinates $f: x \rightarrow f(x)$. We focus here on transformations which can be represented by a smooth field of displacements $\tau$ in the space of coordinates. Equation \ref{eq:taylor} can then be rewritten as follows:
\begin{equation}
    \label{eq:smooth}
    T[f(x)](\epsilon) \approx f(x + \epsilon \tau(x))
\end{equation}
We will refer to such transformations as elastic transformations. We will consider them as a linear approximation of a wide range of complex (camera) transformations.

\begin{figure*}
    \centering
    \includegraphics[width=0.95\linewidth]{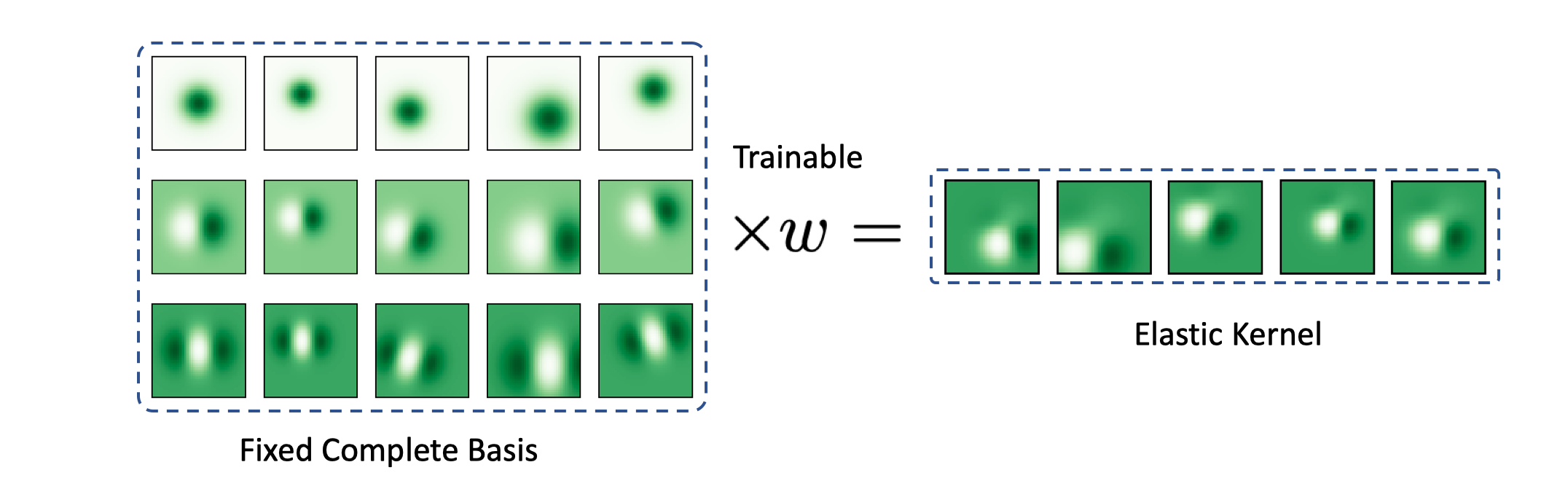}
    \caption{An illustration of how a set of elastic kernels is represented as a trainable linear combination of elastically-augmented fixed basis functions.}
    \label{fig:basis}
\end{figure*}

\subsection{Elastically-Augmented Convolutions}
Let us consider a convolutional layer $\Phi$ parameterized by a filter $\kappa$. It takes input image $f$. The output is:
\begin{equation}
    \label{eq:conv_output}
    \Phi(f, \kappa) = f\star \kappa = \mathbf{K} \times \mathbf{f}
\end{equation}
where $\mathbf{K}$ is a matrix representation of the filter. While, when data augmentation is used, a transformed version of the image can be fed as an input.
\begin{equation}
\begin{split}
    \label{eq:conv_output_transformed}
    \Phi(T[f], \kappa) &= T[f]\star \kappa = \mathbf{K} \times (\mathbf{T}\times\mathbf{f}) 
    = (\mathbf{K} \times \mathbf{T}) \times \mathbf{f} = \Phi(f, T'[\kappa]) 
\end{split}
\end{equation}
In the most general case, $\mathbf{K} \mathbf{T}$ is a matrix representation of a zero padding, followed by a convolution with a kernel and a cropping afterwards. The size of the kernel $T'[\kappa]$ depends on the nature of the transformation $T$. If the transformation if global the kernel can be of a size bigger than the input image. We will consider only the cases when $T'[\kappa]$ is of the same or of a slightly bigger size than the original one.

To incorporate the data augmentation into the  convolutional layers of the network, we propose \textit{elastically-augmented convolutions}, shortly EAConv, as follows:
\begin{equation}
    \text{EAConv} = \max
    \begin{bmatrix}
        \beta_0 \Phi(f, \kappa) \\
        \beta_1 \Phi(f, T_1[\kappa]) \\
        \vdots \\
        \beta_n \Phi(f, T_n[\kappa])
    \end{bmatrix}
\end{equation}
where $\beta_i$ are trainable coefficients. We initialize them such that $\beta_0=1$ and the rest are zeros. The maximum is calculated per pixel among different transformations of the kernel. At the beginning of training, the operation is thus identical to the original convolution with the same filter. If it is required during training, the other coefficients will activate the corresponding transformations.

\subsection{Transformations of a Complete Basis}
In order to apply elastic transformations to filters, we parametrize each filter as a linear combination of basis functions:
\begin{equation}
    \label{eq:filter_basis}
    \kappa = \sum_i w_i \psi_i
\end{equation}
where $\psi_i$ are functions of a complete fixed basis and $w_i$ are trainable parameters. The approach is illustrated in Figure \ref{fig:basis}. We follow \cite{jacobsen2016structured} and choose a basis of 2-dimensional Gaussian derivatives. 

The transformations when applied to the basis form a transformed basis. Thus, for every transformation from the set, there is a corresponding transformed basis. Weights $w_i$ are shared among all bases.

\begin{figure*}
    \centering
    \includegraphics[width=\linewidth]{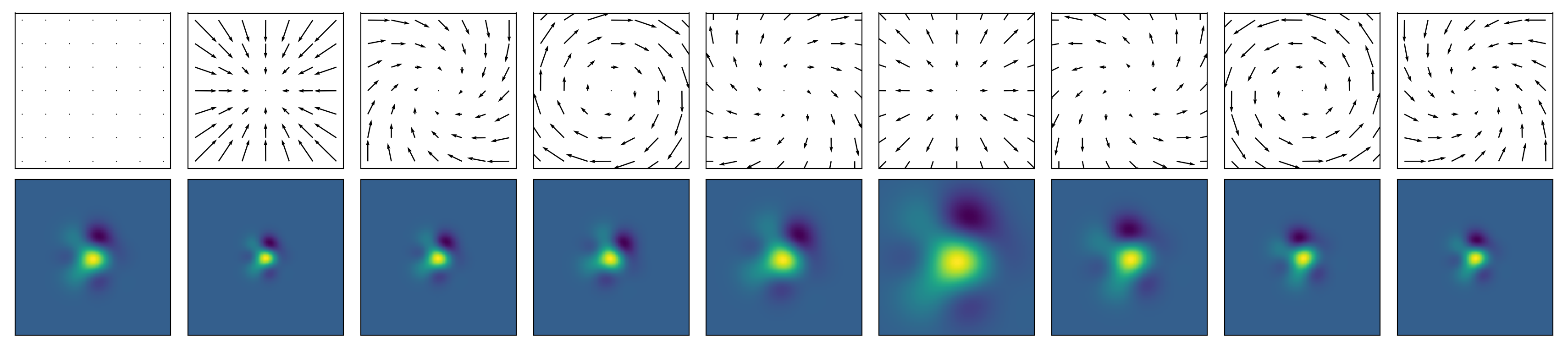}
    \caption{Top: vector fields of smooth displacements for the proposed set of rotation-scaling
transformations. Bottom: the original filter and its versions transformed after applying the
corresponding displacements.}
    \label{fig:kernel_proj_circular}
\end{figure*}

Let us assume that the center of a filter is a point with coordinates $(0, 0)$. For every function from the basis, we first generate a grid of coordinates $(x, y)$. Then we evaluate the value of the function in the coordinates when projected on the pixel grid. In order to transform the functions, we add a small displacement to the coordinates, which leaves the center untransformed. We propose a set of transformations which we call rotations-scaling displacements. See Figure \ref{fig:kernel_proj_circular}. Given a grid of coordinates $(x,y)$, $\alpha$ the elasticity coefficient and $\sigma$ be the scaling factor, we define rotation-scaling displacements as follows:
\begin{align}
\label{eq:rot_scale}
    x'&= x+\alpha(x\text{cos}(\theta)+y\text{sin}(\theta)) \\
    y'&= y+\alpha(-x\text{sin}(\theta)+y\text{cos}(\theta))
\end{align}
where $x',y'$ are the displaced coordinates. And $\theta$ is the scale-rotation parameter. When $\cos(\theta)$ is equal to 0 the whole transformation parametrizes rotation. When $\sin(\theta)$ is equal to 0 then it performs scaling. For all other cases the transformation is a combination of both. The elasticity coefficient contols the severity of the transformations. Thus for the case of rotation it is a linear approximation the $\sin$ of the rotation angle. For the case of scaling, $\alpha$ the scaling coefficient.
% \newpage

We follow \cite{sosnovik2019scale} and use a basis of 2 dimensional hermite polynomials with Gaussian envelope:
\begin{equation}
\label{eq:herm}
    \psi_\sigma(x',y')= A \frac{1}{\sigma^2}H_n\left(\frac{x'}{\sigma}\right)H_m\left(\frac{y'}{\sigma}\right)\text{exp}\left[-\frac{x'^2+y'^2}{2\sigma^2}\right]
\end{equation}
where, $A$ is the normalization constant, $H_n$ is the Hermite polynomial of $n-$th order and $\sigma$ is the scaling factor. We iterate over $n,m$-pairs to generate functions.

\subsection{Elastically-Augmented Residual blocks}
In order to transform residual networks, we propose a straightforward generalization of the proposed convolution. The standard residual block can be formulated as follows:
\begin{equation}
    \text{ResBlock} =  f + G(f, \kappa_1, \kappa_2, \dots)
\end{equation}
The according augmented block is formulated as follows:
\begin{equation}\label{eq:EAResBlock}
    \text{EAResBlock}=  f + \max
    \begin{bmatrix}
        \beta_0 G(f, \kappa_1, \kappa_2, \dots) \\
        \beta_1 G(f, T_1[\kappa_1], T_1[\kappa_2], \dots) \\
        \vdots \\
        \beta_n G(f, T_n[\kappa_1], T_n[\kappa_2], \dots)
    \end{bmatrix}
\end{equation}
Elastic kernels augmented in the network architecture are shown in the Figure.\ref{fig:Architecture}.

\section{Experiments and Results}
We consider two datasets of varying input sizes, i.e. CIFAR-10 $32\times32$ pixels, STL-10 $96\times96$ pixels for our experiments. CIFAR-10 consists of ten coarse-grained classes with 50000 training and 10000 test images \cite{krizhevsky2009learning}. STL-10 contains 5000 training and 8000 test images in ten coarse-grained categories \cite{coates2011analysis}.  

\subsection{Standard Network} We begin by training and testing standard networks for each dataset on clean images. For CIFAR-10, we finetune a Resnet-152 network pretrained on imagenet and achieve $92.53$ on the clean test set.  For STL-10, we train a Wide-Resnet-16 (WRN-16) from scratch and gain $88.28$ on clean images. We also train a Resnet-18 and Resnet-152 for STL-10 pretrained on Image-net and get $83.20$ and $84.10$ clean image accuracy respectively. The only data augmentation used while training is random horizontal flip.

\subsection{Elastically-Augmented Convolutional Network}
Next, we train each classifier network with elastically augmented convolutions. For CIFAR-10, we initialize the weights of EAConv Resnet152 with Imagenets weights and finetune it. While for STL-10, we initialize the weights of the elastically augmented WRN-16 with the weights from a standard network trained on STL-10, and Resnet-18 and Resnet-152 with Imagenet weights. 

\paragraph{Weights Transfer.} In order to train neural networks successfully, initializing neural networks with Imagenet pretrained model weights is a common practice. However, it is not straight forward to transfer the weights of a standard network to our EAConv network because our network is composed of fixed basis and trainable weights, i.e. multiple parallel networks connected to each other Figure \ref{fig:Architecture}. Inspired by Sosnovik \etal \cite{sosnovik2021scale} we assume that in EAConv there is a subnetwork which is identical to the standard network, hence, we can transfer the weights of the standard network to our EAConv subnetwork. 
We start by disconnecting parallel networks by initializing all the weights responsible for inter correlations to zero. Now, the EAConv network until the EAConv max pooling layer (equation \ref{eq:EAResBlock}) consists of several parallel networks disconnected to each other. Convolutional layers of EAConv for which filter sizes match with the standard network, we initialize them with the weights from the standard network. $1\times1$ convolutions of the standard network and the EAConv network are identical, therefore, we copy the weights from the standard to the EAConv network. 
\begin{figure*}
    \centering
    \includegraphics[width=\linewidth, trim=0 0 0 0, clip]{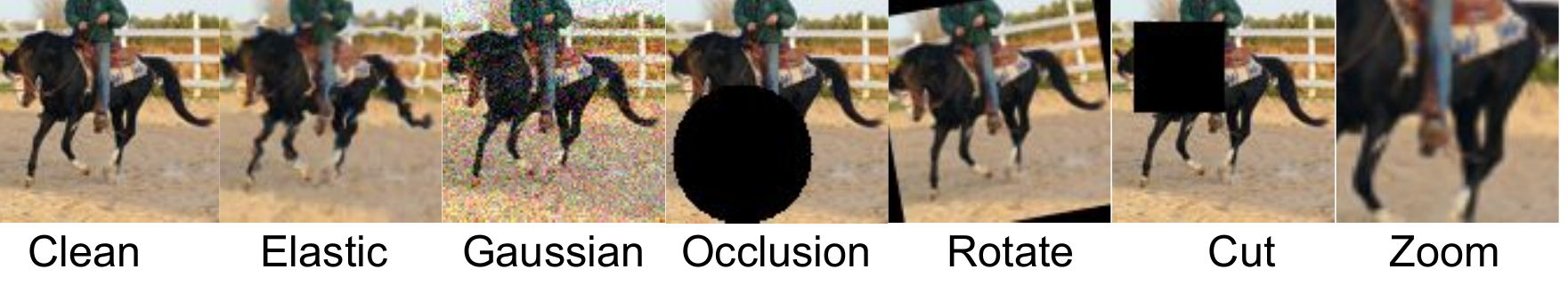}
    \caption{Sample image from STL-10 dataset showing the clean and six different perturbations used in our experiments.}
    \label{fig:sample}
\end{figure*}
\subsubsection{Elastic CIFAR-10.}  For CIFAR-10, we experiment by augmenting  the first convolutional layer and two resnet blocks with elastically augmented convolutions and select the combination which gives the best performance on clean samples.  Introducing EAConv only to the first convolutional layer gives the best performance, i.e. $94.50\%$ for CIFAR-10. Therefore, we select Resnet-152 with the EAConv on the first layer for further experiments. The hyperparameters for EAConv i.e. $\alpha$ and $\sigma$ equation \ref{eq:rot_scale} and \ref{eq:herm} are also selected based on the performance on clean test set. We also search for the best hyperparameters to use in data augmentation.

Table.\ref{table:CIFAR10} contrasts the performance of a standard network, a network with the data augmentation and our EAConv network on CIFAR-10, both for clean and elastic perturbed images. Parameters for perturbations are selected to induce a drop of $5.32\%$ and $10.29\%$. Results show that both our model and data augmentation leads to a recovery in the performance on elastic perturbed images. Although the data augmentation shows better recovery against perturbed inputs, however it does not show any improvement in the performance on clean test set, while our model shows an improvement of $1.97\%$ on clean images. Therefore, elastic data augmentation leads to a bias towards elastic perturbations, while our EAConv show generalization to both perturbed and clean samples.  
\begin{table} 
\begin{center}
\begin{tabular}{|l | c | c | c | }
\hline 
Model & Clean &Drop $\approx 5\%$ &Drop $ \approx 10\%$     \\\hline \hline
Standard Network $(\alpha =0.00,\; \sigma=0.00)$  &$92.53$	&$87.21$	&$82.24$\\ 
Data Augmentation $(\alpha =0.06,\; \sigma=1.28)$ & $92.05$ &	$\textbf{90.89}$&	$\textbf{88.26}$\\
EAConv Network $(\alpha =0.50,\; \sigma=1.00)$ (ours) & $\mathbf{94.50}$ &	$\textbf{90.07}$&	$\textbf{85.55}$\\ \hline
\end{tabular}
\end{center}

\caption{Performance comparison of a standard network and our EAConv Network for CIFAR-10 clean and perturbed inputs. Drop $\approx 5\%$ and $\approx 10\%$ are drops in the performance due to elastic perturbations for a standard network. Although data augmentation shows better recovery in the drop for perturbed images, however our EAConv network besides recovering the drop also improves performance on clean images.  Where $\alpha$ is the elasticity coefficient and $\sigma$ is the scaling factor, eq \ref{eq:rot_scale} and \ref{eq:herm}.}\label{table:CIFAR10}
\end{table} 
\subsubsection{Elastic STL-10.}
For STL-10 WRN-16, we augment all residual blocks with EAConv and test the performance on clean images. We select the augmented combination of layers which gives the best performance on clean images, i.e. $88.93$. Results showed that EAConv at the first convolutional layer, Block0 and Block1, gave us the best performance. For STL-10 Resnet-18 and Resnet-152, we augment EAConv only at the first layer, and it gave us significant improvement in the performance, i.e. $88.49$ for Resnet-18 and $94.48$ for Resnet-152.  Hence, we select WRN-16 with EAConv till Block1, and Resnet-18 and Resnet-152 with EAConv at the first layer for further experiments.

\begin{figure}
        { \includegraphics[width=\linewidth]{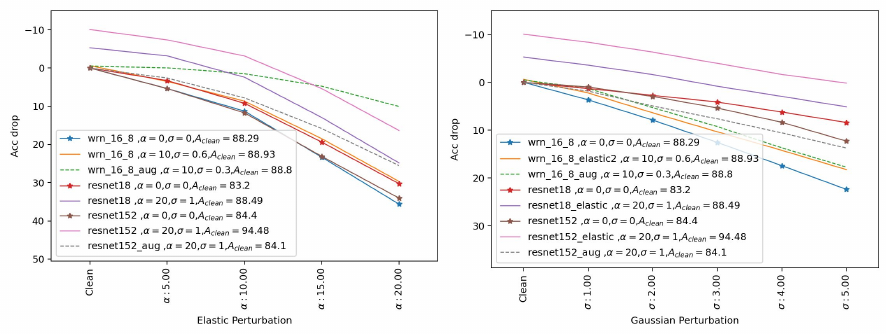}}
         \caption{Evaluating the performance of EAConv on elastic and Gaussian perturbations at different perturbation severity levels on the x-axis. The left plot shows that our method generalizes to elastic perturbations. While, the right plot shows that our method also generalizes to unseen Gaussian perturbations. Resnet-152 performs the best. Where $\alpha$ is the elasticity coefficient, and $\sigma$ is the standard deviation.}
         \label{fig:seen}
\end{figure}

\begin{figure}
         \centering
         \includegraphics[width=12cm]{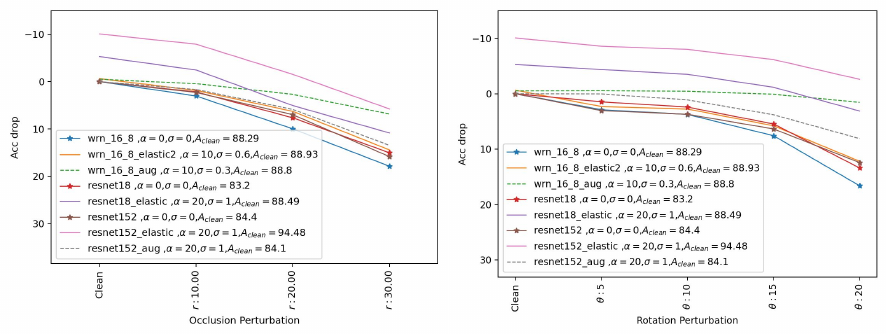}
         \caption{Evaluating the performance of EAConv on unseen occlusion and rotation perturbations at different levels of severity on x-axis. The left plot shows that our method generalizes to occlusions. While, the right plot shows that our method generalizes to rotation perturbations. Where $r$ is the radius of circular occlusions, and $\theta$ is the rotational angle.}
         \label{fig:unseen1}
\end{figure}

\paragraph{Evaluating on Seen Perturbations.} We evaluate the performance of our robustified elastically augmented network on elastic perturbations in Figure.\ref{fig:seen} (left). In the plot on the x-axis, we have a clean test set and four elastically perturbed test sets with varying severity levels. While on the y-axis, we have the drop in the accuracy with the clean test on a standard network with the drop zero and increasing drop with the increase in the severity levels. Solid lines with the star symbol show the drop on a standard WRN-16, Resnet-18 and Resnet-152 for elastically perturbed samples. Solid lines without the star symbol show the performance of our EAConv rotation scaling transforms, and the dotted lines depict a standard network trained with elastic data augmentation. 

 Results show that with the increase in the perturbations the accuracy drops, however, our EAConv network recovers the drop for all the severity levels while enhancing the performance on clean test set. We observe that Resnet-18 and Resnet-152 show significant recovery as compared to WRN-16, this is because WRN-16 \cite{zagoruyko2016wide} is designed to have less depth but more width, and it saturates at $88.93$ (without any data augmentation), therefore it shows a lack in capacity to capture view point variations.
 
 Although data augmentation helps against elastic perturbations with WRN-16, but it does not help with Resnet-18 and Resnet-152. Additionally, the clean image performance stays the same. Hence, WRN-16 data augmentation does not improve performance for clean images, however it generalizes to elastic perturbation. Our EAConv generalizes to the perturbations while improving performance on clean images leading to state of the art with Resnet-152 EAConv i.e. $94.48$.
\begin{figure}
         \centering
         \includegraphics[width=12cm]{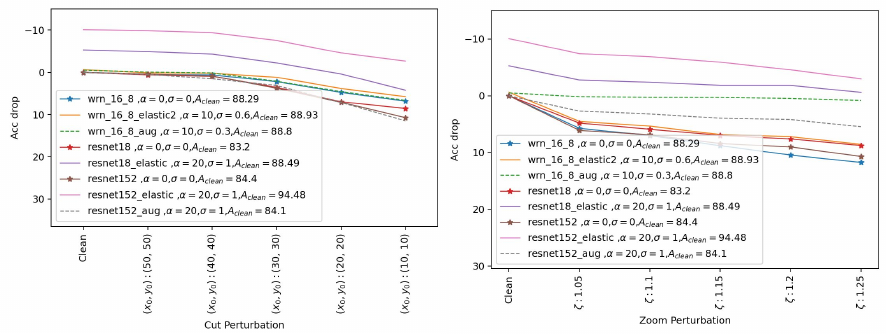}
         \caption{Evaluating the performance of EAConv on unseen cut and zoom perturbations at different levels of severity on x-axis. The left plot shows that our method generalizes to cuts in the images. While, the right plot shows that our method generalizes to zoom perturbations. Where location is the starting point in pixels for the cut, e.g. $(x_0,y_0)=(50,50)$, and $\zeta$ is the zoom factor. }
         \label{fig:unseen2}
\end{figure}
\paragraph{Evaluating on Unseen Gaussian Perturbations.}
 Figure.\ref{fig:seen} (right) shows the effectiveness of our method on unseen Gaussian perturbations. On the x-axis, we have a clean test set and five different test sets with the Gaussian perturbations of varying standard deviation $std=1$ to $5$. The plots show that with the increase in the severity of Gaussian noise, the accuracy drops for the standard networks (solid lines with star symbol). However, when we test our EAConv networks (solid lines without star symbol) on these perturbations, it helps to recover the drop, Resnet-152 EAConv being the best one. In contrast, data augmentation with elastic deformations show small improvement in the performance for WRN-16 and leads to a further drop for Resnet-152 (dotted lines).

\paragraph{Evaluating on Unseen Occlusion Perturbations.}
Figure \ref{fig:unseen1} (left) shows the performance of networks tested on a clean and three different occluded test sets with varying sizes of occlusion from radius $r=$ $10$ to $50$. Occlusion is a circle with the radius $r$ and the center of the circle is selected randomly between $r$ and the image size. Hence, the position of the occlusion varies for each image. We observe that the classification accuracy drops with the increase in the size of occlusion on a standard networks (solid lines with star symbol). However, our EAConv network (solid lines without star symbol) shows recovery in the drop, hence generalizing to unseen occlusion perturbations. Data augmentation with elastic perturbations WRN-16 shows robustness for large occlusions, but it is less than EAConv Resnet-152 (dotted lines).

\paragraph{Evaluating on Unseen Rotation Perturbations.}
The plot in Figure \ref{fig:unseen1} (right) contrasts the performance of standard networks with EAConv networks and data augmentations for varying rotation perturbations with the angle $\theta = 5$ to $20$. Results show that rotating images lead to a drop in the performance for standard networks (solid lines with star symbol), however our EAConv networks generalizes to these rotation perturbations (solid lines without star symbol). Data augmentation with WRN-16 also shows recovery in the performance, however it is less than our EAConv Resnet-152 (dotted lines).   

\paragraph{Evaluating on Unseen Cut Perturbations.}
 For cut perturbation Figure \ref{fig:unseen2} (left), we cut the part of the image equal to half the image size starting at the location on x-axis in the plot, e.g. $(x_0,y_0)=(50,50)$. As the location changes towards the beginning pixels of the image, a large part of the image is cut out. Hence, leading to higher drop in the performance (solid lines with star symbol).  Our EAConv with Resnet-18 and Resnet-152 recovers the drop significantly, WRN-16 shows improvement for large drops (solid lines without star symbol). Data augmentation does not help to recover the drop. 

\paragraph{Evaluating on Unseen Zoom Perturbations.}
Finally, Figure \ref{fig:unseen2} (right) shows the performance of the standard, EAConv and data augmentation in the presence of zoom perturbations. On the x-axis we vary the zoom factor $\zeta$, we observe that with the increase in the zoom the classification accuracy drops (solid lines with star symbol). Our EAConv Resnet-18 and Resnet-152 help to recover the drop significantly. Data augmentation also shows generalization to zoom, however it is less than our EAConv networks.

Hence, our EAConv network helps to improve the general robustness on unseen perturbations without any extra cost of data augmentation.

%--------------------------------------------------------

\section{Conclusion}
A method to integrate unseen view points in the convolutional neural networks is introduced for enhancing the robustness against local variations in the image space. We demonstrated the effectiveness of our method by improving the performance on perturbed test inputs while enhancing the generalization on clean test inputs. We also showed general robustness of our EAConv network by testing on unseen occlusion, cut, zoom, rotation and Gaussian perturbations. Our results showed that elastically augmented convolutions enhance the robustness against unseen viewpoint variations while keeping the number of training parameters in the network and the number of training images the same. Moreover, it improves the accuracy on clean images for both CIFAR-10 and STL-10 datasets, reaching the state of the art without any data augmentation.
\bibliography{egbib}
\end{document}